\documentclass[12pt,english]{article}
\usepackage[T1]{fontenc}
\usepackage[latin9]{inputenc}
\usepackage[letterpaper]{geometry}
\usepackage{array}
\usepackage{amssymb}

\makeatletter

\providecommand{\tabularnewline}{\\}

\makeatother

\usepackage{babel}

\begin{document}

\title{Toward Measuring the Scaling of Genetic Programming}

\author{Mike Stimpson}
\maketitle
\begin{abstract}
\noindent Several genetic programming systems are created, each solving
a different problem. In these systems, the median number of generations
$G$ needed to evolve a working program is measured. The behavior
of $G$ is observed as the difficulty of the problem is increased.

\noindent In these systems, the density $D$ of working programs in
the universe of all possible programs is measured. The relationship
$G\sim\frac{1}{\sqrt{D}}$ is observed to approximately hold for two
\emph{program-like} systems.

\noindent For \emph{parallel} systems (systems that look like several
independent programs evolving in parallel), the relation $G\sim n\ln n$
is observed to approximately hold.

\noindent Finally, systems that are \emph{anti-parallel} are considered. 
\end{abstract}

\section{INTRODUCTION}

\noindent Most genetic programming experiments appear to evolve solutions
to very small problems - small in terms of program size and/or the
number of variables used. For example, people have evolved sorting
programs. Nobody has evolved an operating system, a database, or a
working air-traffic control system for the United States.

\noindent It seems, then, that genetic programming doesn't scale well
to larger, more difficult problems. But how do you measure how large
or how difficult a problem is?

\noindent My initial approach was to keep the problem constant, and
vary the set of statements that programs were implemented in. (It
is a common observation that it is more difficult to write programs
in a low-level language than in a high-level one. That is, the same
problem is {}``harder'' or {}``more complicated'' when written
in a low-level language, and the source code for the program to solve
the problem is larger.) The difference in how many generations it
took to evolve a working program would then be entirely due to the
change of difficulty of implementing the program in the statement
set, because the problem would be constant. This approach was used
for the first two systems. As research progressed, it became apparent
that other, sometimes system-specific parameters gave more precise
control over the difficulty of the problem.

\noindent The rest of this paper is organized as follows: Section
2 describes the first system. Section 3 presents the results in the
form of a number of datasets, each of which contains only one varying
parameter. Section 4 demonstrates the relationship between the density
of working programs and the median number of generations needed to
evolve a working program. Section 5 describes a second system that
exhibits similar scaling behavior. Section 6 describes two parallel
systems - systems that have multiple dimensions, where each dimension
can be optimized independently. Section 7 describes three anti-parallel
systems - systems that have multiple dimensions, but none of the dimensions
can be optimized independently. Section 8 presents some conclusions,
and section 9 presents some open questions.

\section{\noindent THE FIRST SYSTEM: LINEAR PROGRAMS, SORTING INTEGERS}

\noindent I chose sorting a list of integers as the first problem.

\noindent This system had a fixed number $v$ of writable variables,
numbered 1 through $v$. ({}``Fixed'' here means that it did not
evolve; however, it could be changed between runs via a command-line
parameter.) It also contained two read-only {}``variables''. Variable
0 always contained 0, and variable $v+1$ always contained the number
of integers in the list being sorted.

\noindent A program consisted of a series of statements. Within one
run, all programs of all generations had the same length.

\noindent The initial programs contained random statements. The default
population size was 20 programs. The most fit programs (default 4)
were chosen to produce the next generation. If there was a tie among
programs for which was most fit (or, more importantly, 4th-most-fit),
a winner was randomly selected from the tied programs.

\noindent Fitness was tested by having each program attempt to sort
three lists of numbers, which respectively contained 10, 30 and 50
values. The lists contained the values from 1 to the size of the list,
in random order. After a program attempted to sort a list, the \emph{forward
distance} was computed as follows: For each location in the list,
the absolute value was taken of the difference between the value at
that location in the list as sorted by the program, and the value
that would be at that location if the list were perfectly sorted.
A perfectly sorted list therefore had a forward distance of zero.
The \emph{reverse distance} was identical, except that the {}``perfectly
sorted'' list was replaced by one that was perfectly sorted in reverse
order. In general, the forward and backward distances were larger
for the longer lists. To address this, a \emph{normalized metric}
was created for each list, which was the reverse distance minus the
forward distance, divided by the sum of the forward and reverse distances.
This evaluated to 1 for a perfectly sorted list, and to $-1$ for
a list that was perfectly sorted in reverse. Finally, the program's
fitness function was the average of the normalized metrics for the
three lists.

\noindent A program was considered to be terminated when the last
statement was executed, if the last statement was not a jump, or when
a jump was executed to one past the last statement. (This is equivalent
to saying that all programs had an \texttt{End} as the assumed last
statement, and the \texttt{End} could not mutate.) If the program
executed 10 times as many statements as were required for a bubble
sort for the same list, the program was considered to be in an infinite
loop, and terminated. No fitness penalty was imposed for this condition.

\noindent Cross-breeding was done by choosing two programs, randomly
choosing a location within the list of statements of the programs
(the same location for both), cutting each program into two pieces
at that location, and swapping the pieces to create two child programs.
This ensured that the children were the same length as the parents.
Also, during this process, a statement in a child program could randomly
mutate into another statement with some probability (default 0.2).

\noindent Programs were composed of statements (instructions) that
were members of a set of statements.

\noindent Statement set 1 contained two statements: \texttt{CompareSwap}
(compare two numbers in the list, and swap them if they are out of
order), and \texttt{For} (a C-style for loop with a loop variable,
a variable from which to initialize the loop variable, and a limit
variable to compare the loop variable to). Programs with this statement
set defaulted to 5 statements long, even though a bubble sort can
be written with three such statements (two \texttt{For} statements
and one \texttt{CompareSwap} statement). This {}``slack'' in the
number of statements gave more rapid evolution than a length that
had no more statements than were absolutely necessary.

\noindent Statement set 2 contained \texttt{IfVarLess} (if the value
in one register is less than the value in another register, execute
the next statement), \texttt{IncrementVar} (increment a register),
\texttt{AssignVar} (copy the value from one register to another),
\texttt{GoTo}, and \texttt{CompareSwap}. With this statement set,
I could write a bubble sort in 11 statements, but programs created
from statement set 2 evolved better with 25 statements per program.

\noindent Statement set 3 contained \texttt{IfVarLess}, \texttt{IncrementVar},
\texttt{AssignVar}, \texttt{GoTo}, \texttt{IfListLess }(if the list
entry at the index contained in the first variable is less than the
list entry at the index contained in the second variable, execute
the next statement), and \texttt{Swap} (an unconditional swap). With
this statement set, I could write a bubble sort in 12 statements,
but programs created from statement set 3 evolved better with 30 statements
per program.

\noindent The unsorted lists of numbers were randomly created. New
lists were created for each generation. (If the same lists were used
for all generations, statement set 2 would sometimes be unable to
evolve a working program.)

\noindent An \emph{evolution} started with a random collection of
programs, and proceeded until a program evolved that worked. An evolution
was characterized by the number of generations required to evolve
a working program. However, since evolution is a random process, a
repeat of the evolution would take a completely different number of
generations.

\noindent A \emph{run} was a number of evolutions, all with the same
statement set and the same parameters. It was characterized by the
median of the number of generations required for each evolution in
the run. (The distribution of the number of generations had a very
long tail. The presence or absence of one anomalous evolution could
significantly shift the average, so the median was the appropriate
choice here.)

\noindent For statement set 1, runs consisted of 1000 evolutions,
and the results were quite repeatable (within 10\%, and often closer
to 2\%). For statement sets 2 and 3, runs were reduced to 100 evolutions,
because the evolutions took far longer (both because it took more
generations to evolve a solution, and because each program could execute
many more statements before it was declared to be in an infinite loop).
Re-runs of statement sets 2 and 3 could give results that differ by
as much as 30\% from the first run.

\noindent I also measured the density of working programs in the universe
of all possible programs, by generating a large number of random programs
and seeing how many of them worked {}``as is'', that is, with no
evolution. When measuring density for statement set 1, I made sure
that the sample was large enough to contain at least 1000 working
programs. For statement sets 2 and 3, I only tried for a sample large
enough to contain 100 working programs, because otherwise the density
run times became extremely long.

\section{DATA AND ANALYSIS}

\subsection*{Changing number of variables}

Statement set 1:

\begin{tabular}{|c|c|}
\hline 
Number of variables & Median generations\tabularnewline
\hline 
2 & 242\tabularnewline
\hline 
3 & 341\tabularnewline
\hline 
4 & 450\tabularnewline
\hline 
5 & 771.5\tabularnewline
\hline 
7 & 1238.5\tabularnewline
\hline 
10 & 2674.5\tabularnewline
\hline 
20 & 10580\tabularnewline
\hline
\end{tabular}

Adding variables increased the size of the solution space. As the
solution space got larger (and the number of programs that work increased,
too, but not as fast, as we will see in the next section), the number
of generations climbed dramatically.

Statement set 2:

\begin{tabular}{|c|c|}
\hline 
Number of variables & Median generations\tabularnewline
\hline 
2 & 30101\tabularnewline
\hline 
3 & 39742.5\tabularnewline
\hline 
5 & 51648.5\tabularnewline
\hline 
7 & 167621.5\tabularnewline
\hline 
10 & 152464\tabularnewline
\hline
\end{tabular}

As the statements became simpler, the problem became more complex
in terms of the solution language, and the number of generations exploded.
(Statement set 2 takes 11 statements to write a bubble sort in, versus
3 statements for statement set 1, that is, statement set 2 takes 3.67
times as many statements to implement one particular algorithm to
solve this problem. But it took 124 times as many generations to evolve
a working program with 2 variables, and 57 times as many generations
to evolve one with 10 variables.)

\pagebreak{}Statement set 3:

\begin{tabular}{|c|c|}
\hline 
Number of variables & Median generations\tabularnewline
\hline 
2 & 34934\tabularnewline
\hline 
3 & 53008\tabularnewline
\hline 
5 & 58697.5\tabularnewline
\hline 
7 & 83301.5\tabularnewline
\hline 
10 & 83301.5\tabularnewline
\hline
\end{tabular}

This is quite surprising! Even though the problem grew more complex
in terms of the statements, the median number of generations went
down dramatically, especially with a larger number of variables.

A possible explanation would be the program length. (Statement set
3 defaulted to 30 statements per program, and statement set 2 to 25.)
But with 10 variables and 25 statements per program, statement set
2 took 152464 generations, and statement set 3 took 115857. So program
length doesn't seem to be the reason that statement set 2 took more
generations than statement set 3.

I see another possible explanation, however: In statement set 2, it
was hard to build a loop - it took 5 statements. But one \texttt{CompareSwap}
could give you, on average, \emph{some} improvement. So a mutation
from some other statement to a \texttt{CompareSwap} statement could
destroy a working (or almost working) loop and actually improve the
program's score. Statement set 1 didn't have this problem, since loops
were only one statement long. Statement set 3 didn't have this problem,
either, since an unconditional \texttt{Swap}, on average, would not
cause any improvement.

This may not be the correct explanation of this anomaly. But we are
going to see in the next section that \emph{something} is very wrong
with statement set 2.

Since statement set 2 is somewhat suspect, let us repeat the previous
comparison with statement set 3. Statement set 3 takes 12 statements
to write a bubble sort in, versus 3 statements for statement set 1,
so statement set 3 takes 4 times as many statements to implement one
particular algorithm to solve this problem. But it took 144 times
as many generations to evolve a working program with 2 variables,
and 48 times as many generations to evolve one with 10 variables.

\subsection*{Changing Population Size (number of programs and number of parents)}

Statement set 1, 2 variables:

\begin{tabular}{|c|c|c|}
\hline 
Number of programs & Number of parents & Median generations\tabularnewline
\hline 
20 & 4 & 242\tabularnewline
\hline 
40 & 8 & 90.5\tabularnewline
\hline 
80 & 16 & 18\tabularnewline
\hline
\end{tabular}

\pagebreak{}Statement set 1, 5 variables:

\begin{tabular}{|c|c|c|}
\hline 
Number of programs & Number of parents & Median generations\tabularnewline
\hline 
20 & 4 & 771.5\tabularnewline
\hline 
40 & 8 & 384\tabularnewline
\hline 
80 & 16 & 156.5\tabularnewline
\hline 
160 & 32 & 25\tabularnewline
\hline
\end{tabular}

For statement set 1, it seems that doubling the number of programs
(and parents) halved the number of generations, as long as the number
of generations didn't get too small. When the number of generations
got below about 100 or 200, doubling the number of programs resulted
in less than half of the number of generations.

Statement set 2, 2 variables:

\begin{tabular}{|c|c|c|}
\hline 
Number of programs & Number of parents & Median generations\tabularnewline
\hline 
20 & 4 & 33025\tabularnewline
\hline 
40 & 8 & 29454.5\tabularnewline
\hline 
80 & 16 & 18173\tabularnewline
\hline 
160 & 32 & 17886\tabularnewline
\hline 
320 & 64 & 7580\tabularnewline
\hline 
640 & 128 & 7062.5\tabularnewline
\hline 
1280 & 256 & 5544\tabularnewline
\hline
\end{tabular}

Doubling the number of programs in statement set 2 seemed to give
much \emph{less} than 50\% improvement in the number of generations.
The only exception was going from 160 programs to 320, where the number
of generations was reduced by 58\%.

\subsection*{Changing Program Length}

Statement set 1, 2 variables:

\begin{tabular}{|c|c|}
\hline 
Number of statements & Median generations\tabularnewline
\hline 
3 & 331.5\tabularnewline
\hline 
4 & 271\tabularnewline
\hline 
5 & 242\tabularnewline
\hline 
6 & 240\tabularnewline
\hline 
8 & 267.5\tabularnewline
\hline 
10 & 313.5\tabularnewline
\hline 
12 & 278.5\tabularnewline
\hline 
15 & 302\tabularnewline
\hline 
20 & 368\tabularnewline
\hline
\end{tabular}

\pagebreak{}Statement set 1, 3 variables:

\begin{tabular}{|c|c|}
\hline 
Number of statements & Median generations\tabularnewline
\hline 
3 & 425\tabularnewline
\hline 
4 & 402.5\tabularnewline
\hline 
5 & 341\tabularnewline
\hline 
6 & 282.5\tabularnewline
\hline 
8 & 312\tabularnewline
\hline 
10 & 306.5\tabularnewline
\hline 
12 & 352.5\tabularnewline
\hline 
15 & 374\tabularnewline
\hline 
20 & 390.5\tabularnewline
\hline
\end{tabular}

Statement set 1, 5 variables:

\begin{tabular}{|c|c|}
\hline 
Number of statements & Median generations\tabularnewline
\hline 
3 & 976\tabularnewline
\hline 
4 & 819\tabularnewline
\hline 
5 & 771.5\tabularnewline
\hline 
6 & 626.5\tabularnewline
\hline 
8 & 525\tabularnewline
\hline 
10 & 557.5\tabularnewline
\hline 
12 & 604\tabularnewline
\hline 
15 & 552.5\tabularnewline
\hline 
20 & 614\tabularnewline
\hline
\end{tabular}

Statement set 1, 7 variables:

\begin{tabular}{|c|c|}
\hline 
Number of statements & Median generations\tabularnewline
\hline 
3 & 1654.5\tabularnewline
\hline 
4 & 1448\tabularnewline
\hline 
5 & 1238.5\tabularnewline
\hline 
6 & 1070\tabularnewline
\hline 
8 & 891\tabularnewline
\hline 
10 & 880.5\tabularnewline
\hline 
12 & 799.5\tabularnewline
\hline 
15 & 852\tabularnewline
\hline 
20 & 921\tabularnewline
\hline
\end{tabular}

\pagebreak{}Statement set 1, 10 variables:

\begin{tabular}{|c|c|}
\hline 
Number of statements & Median generations\tabularnewline
\hline 
3 & 3598.5\tabularnewline
\hline 
4 & 3032.5\tabularnewline
\hline 
5 & 2674.5\tabularnewline
\hline 
6 & 2057\tabularnewline
\hline 
8 & 1774\tabularnewline
\hline 
10 & 1569.5\tabularnewline
\hline 
12 & 1714\tabularnewline
\hline 
15 & 1677\tabularnewline
\hline 
20 & 1628.5\tabularnewline
\hline
\end{tabular}

Statement set 1, 12 variables:

\begin{tabular}{|c|c|}
\hline 
Number of statements & Median generations\tabularnewline
\hline 
3 & 5142.5\tabularnewline
\hline 
4 & 4207.5\tabularnewline
\hline 
5 & 3480.5\tabularnewline
\hline 
6 & 3050\tabularnewline
\hline 
8 & 2730\tabularnewline
\hline 
10 & 2378\tabularnewline
\hline 
12 & 2351.5\tabularnewline
\hline 
15 & 2278.5\tabularnewline
\hline 
20 & 2219\tabularnewline
\hline
\end{tabular}

Here we see that the optimal length of the program increased slowly
as the number of variables increased.

\section{RELATIONSHIP BETWEEN SOLUTION DENSITY AND NUMBER OF GENERATIONS}

By \emph{density}, we mean the fraction of working programs within
the universe of all possible programs for that statement set and number
of variables.

Density data for statement set 1:

\begin{tabular}{|c|c|}
\hline 
Variables & Density\tabularnewline
\hline 
2 & $1.313\times10^{-3}$\tabularnewline
\hline 
3 & $6.78\times10^{-4}$\tabularnewline
\hline 
5 & $1.911\times10^{-4}$\tabularnewline
\hline 
7 & $7.54\times10^{-5}$\tabularnewline
\hline 
10 & $2.3\times10^{-5}$\tabularnewline
\hline 
20 & $2.11\times10^{-6}$\tabularnewline
\hline
\end{tabular}

\pagebreak{}Statement set 2:

\begin{tabular}{|c|c|}
\hline 
Variables & Density\tabularnewline
\hline 
2 & $2.59\times10^{-6}$\tabularnewline
\hline 
3 & $1.683\times10^{-6}$\tabularnewline
\hline 
5 & $1.033\times10^{-6}$\tabularnewline
\hline 
7 & $4.96\times10^{-7}$\tabularnewline
\hline 
10 & $1.967\times10^{-7}$\tabularnewline
\hline
\end{tabular}

Statement set 3:

\begin{tabular}{|c|c|}
\hline 
Variables & Density\tabularnewline
\hline 
2 & $3.23\times10^{-8}$\tabularnewline
\hline 
3 & $1.867\times10^{-8}$\tabularnewline
\hline
\end{tabular}

(All of the above densities were with the default number of programs,
and with the default program length for the statement set.)

Combining these densities with the median number of generations to
reach a working program, we observe a pattern: When we hold everything
else constant and change the number of variables, the median number
of generations needed to evolve a working program is almost proportional
to the reciprocal of the square root of the density. That is, if $G$
is the median number of generations and $D$ is the density of working
programs, then $K=G\times\sqrt{D}$ is almost constant. This value
($K$) rises slowly as the number of generations increases and the
density decreases.

Statement set 1:

\begin{tabular}{|c|c|c|c|}
\hline 
Variables & $G$ & $D$ & $K$\tabularnewline
\hline 
2 & 242 & $1.313\times10^{-3}$ & 8.77\tabularnewline
\hline 
3 & 341 & $6.78\times10^{-4}$ & 8.88\tabularnewline
\hline 
5 & 771.5 & $1.911\times10^{-4}$ & 10.67\tabularnewline
\hline 
7 & 1238.5 & $7.54\times10^{-5}$ & 10.75\tabularnewline
\hline 
10 & 2674.5 & $2.3\times10^{-5}$ & 12.84\tabularnewline
\hline 
20 & 10580 & $2.11\times10^{-6}$ & 15.37\tabularnewline
\hline
\end{tabular}

Statement set 2:

\begin{tabular}{|c|c|c|c|}
\hline 
Variables & $G$ & $D$ & $K$\tabularnewline
\hline 
2 & 30101 & $2.59\times10^{-6}$ & 48.4\tabularnewline
\hline 
3 & 39742.5 & $1.683\times10^{-6}$ & 51.6\tabularnewline
\hline 
5 & 51648.5 & $1.033\times10^{-6}$ & 52.5\tabularnewline
\hline 
7 & 167621.5 & $4.96\times10^{-7}$ & 118.1\tabularnewline
\hline 
10 & 152464 & $1.967\times10^{-7}$ & 67.6\tabularnewline
\hline
\end{tabular}

Statement set 3:

\begin{tabular}{|c|c|c|c|}
\hline 
Variables & $G$ & $D$ & $K$\tabularnewline
\hline 
2 & 34934 & $3.23\times10^{-8}$ & 6.28\tabularnewline
\hline 
3 & 53008 & $1.867\times10^{-8}$ & 7.24\tabularnewline
\hline
\end{tabular}

Note how high the $K$ values are for statement set 2 compared to
either statement sets 1 or 3. This is why I said that something was
wrong with statement set 2.

Statement set 2 didn't completely follow the pattern. Looking more
closely, we see an anomaly: 7 variables required more generations
than 10 variables did. I reran both the evolution runs and the densities,
and extended it to 12 variables. The new results were:

\begin{tabular}{|c|c|c|c|}
\hline 
Variables & $G$ & $D$ & $K$\tabularnewline
\hline 
2 & 34501.5 & $2.35\times10^{-6}$ & 52.9 \tabularnewline
\hline 
3 & 56712.5 & $1.917\times10^{-6}$ & 78.5\tabularnewline
\hline 
5 & 68639.5 & $8.3\times10^{-7}$ & 62.5\tabularnewline
\hline 
7 & 139623.5 & $4.87\times10^{-7}$ & 97.4\tabularnewline
\hline 
10 & 194577.5 & $1.85\times10^{-7}$ & 83.7\tabularnewline
\hline 
12 & 255380 & $1.31\times10^{-7}$ & 92.4\tabularnewline
\hline
\end{tabular}

In the repeated run, the anomaly is gone, and the regularity we observed
before is seen to approximately hold.

However, the same regularity did not hold for changing the program
length. Here, though the density actually increased as the program
length increased, the number of generations increased anyway.

Statement set 1, 2 variables:

\begin{tabular}{|c|c|c|c|}
\hline 
Statements & $G$ & $D$ & $K$\tabularnewline
\hline 
3 & 331.5 & $5.7\times10^{-4}$ & 7.91\tabularnewline
\hline 
4 & 271 & $8.86\times10^{-4}$ & 8.07\tabularnewline
\hline 
5 & 242 & $1.313\times10^{-3}$ & 8.77\tabularnewline
\hline 
6 & 240 & $1.525\times10^{-3}$ & 9.37\tabularnewline
\hline 
10 & 313.5 & $2.43\times10^{-3}$ & 15.45\tabularnewline
\hline 
20 & 368 & $3.74\times10^{-3}$ & 22.5\tabularnewline
\hline
\end{tabular}

Statement set 1, 7 variables:

\begin{tabular}{|c|c|c|c|}
\hline 
Statements & $G$ & $D$ & $K$\tabularnewline
\hline 
3 & 1654.5 & $3.7\times10^{-5}$ & 10.06\tabularnewline
\hline 
4 & 1448 & $5.5\times10^{-5}$ & 10.74\tabularnewline
\hline 
5 & 1238.5 & $7.54\times10^{-5}$ & 10.75\tabularnewline
\hline 
6 & 1070 & $9.52\times10^{-5}$ & 10.44\tabularnewline
\hline 
8 & 891 & $1.339\times10^{-4}$ & 10.31\tabularnewline
\hline 
10 & 880.5 & $1.741\times10^{-4}$ & 11.62\tabularnewline
\hline 
12 & 799.5 & $2.19\times10^{-4}$ & 11.83\tabularnewline
\hline 
15 & 852 & $2.72\times10^{-4}$ & 14.03\tabularnewline
\hline 
20 & 921 & $3.59\times10^{-4}$ & 17.46\tabularnewline
\hline
\end{tabular}

It appears, then, that we can say that $K=G\times\sqrt{D}$ is \emph{at
best} almost constant, but the number of generations could be considerably
higher if the program length was not optimal.

Earlier, we saw that the optimal length of program for statement set
1 increased as the number of variables increased. What happens if,
for each number of variables, we take the optimal length?

\begin{tabular}{|c|c|c|c|c|}
\hline 
Variables & Statements & G & $D$ & $K$\tabularnewline
\hline 
2 & 6 & 240 & $1.493\times10^{-3}$ & 9.273\tabularnewline
\hline 
3 & 6 & 282.5 & $8.195\times10^{-4}$ & 8.087\tabularnewline
\hline 
5 & 8 & 525 & $3.472\times10^{-4}$ & 9.782\tabularnewline
\hline 
7 & 12 & 799.5 & $2.188\times10^{-4}$ & 11.826\tabularnewline
\hline 
10 & 10 & 1569.5 & $5.643\times10^{-5}$ & 11.79\tabularnewline
\hline 
12 & 20 & 2219 & $7.003\times10^{-5}$ & 18.57\tabularnewline
\hline
\end{tabular}

\section{THE SECOND SYSTEM: TREE-STRUCTURED PROGRAMS, $n$-BIT PARITY}

For the second system, I changed nearly everything. Instead of sorting
integers, I changed the problem to calculating $n$-bit parity. The
fitness function was, out of all possible inputs of $n$ bits, the
number of inputs for which the program computed the correct parity.
Rather than demanding perfection, a program was regarded as working
if the fitness function equaled or exceeded a threshold value (called
the \emph{termination condition}). Each parent was chosen by randomly
selecting seven programs and choosing the most-fit from among the
seven to be the parent. (This approach was also used for all subsequent
systems.) The mutation probability was decreased to 1\% (also for
all subsequent systems). Instead of a linear list of statements, a
program was represented as a LISP-like tree structure. Programs were
variable-length rather than fixed-length. Mutations could alter a
whole sub-tree, rather than a single node. The default population
size was 1000, rather than 20. And, optionally, subroutines could
be automatically generated. (This is similar to the system in \cite{Koza}.
I also used the same probabilities of creating subroutines and other
program-transforming events as his system used.)

The difficulty was changed by increasing the number of bits, and by
increasing the termination condition.

This system presented a new problem when measuring densities, because
the universe of all possible programs was not a simple $n$-dimensional
cube as it was in the first system. Instead, due to the variable length
of the programs, and the fact that almost all non-leaf nodes took
two arguments, there were about twice as many possible programs with
200 nodes (the maximum length for the system) as there were possible
programs with 199 nodes. In turn, there were twice as many possible
programs with 199 nodes as there were with 198 nodes, and so on. In
fact, about half of the programs in the universe of all possible programs
had the maximum length.

But the evolved programs had a very different length distribution,
with nothing below a length of about 10, then a relatively uniform
distribution up to about 50 nodes, then slowly tailing off, with only
about 10\% (range 0\% to about 40\%) having a length greater than
100 nodes. As noted in \cite{Koza}, programs of exceptional length
rarely contribute much to the solution of a genetic programming problem.
In fact, the longer programs had a lower density of working programs
than shorter programs did.

As an evolution proceeds, the length distribution of the population
of programs should become more and more similar to the distribution
of working programs, and less and less similar to the distribution
of the universe of all possible programs. Given, then, that the universe
of all possible programs is structurally very different from both
the working programs that are evolved and from the population during
an evolution, how can we get meaningful density data? I chose the
approach of trying to create self-consistent population distributions
- that is, population distributions such that, when populations with
that length distribution were evolved, the resulting working programs
had the same distribution of lengths. (In practice, this could only
be approximately achieved.) If we measure the density of a population
of programs with the same length distribution as the working programs,
we obtain density data that we can meaningfully combine with the median
number of generations, to see if the relationship observed with the
first system also holds here. (The alternative - the density data
coming from populations that are unlike the population of working
programs - clearly is less likely to provide meaningful data.) The
same approach - finding self-consistent distributions - was also applied
to the number of subroutines, when subroutines were allowed.

\subsection*{Statement set 1}

Statement set 1 contained the following node types: \texttt{Xor},
\texttt{And}, \texttt{Or}, and \texttt{Not}. Also, there were constant
nodes, which contained either 0 or 1. The results for this statement
set were:

4 bits, no subroutines:

\begin{tabular}{|c|c|c|c|}
\hline 
Termination condition & $G$ & $D$ & $K$\tabularnewline
\hline 
14 & 4 & $3.24\times10^{-4}$ & 0.072\tabularnewline
\hline 
16 & 5 & $2.1\times10^{-4}$ & 0.0725\tabularnewline
\hline
\end{tabular}

5 bits, no subroutines:

\begin{tabular}{|c|c|c|c|}
\hline 
Termination condition & $G$ & $D$ & $K$\tabularnewline
\hline 
24 & 3 & $4.48\times10^{-4}$ & 0.0635\tabularnewline
\hline 
28 & 12 & $3.75\times10^{-5}$ & 0.0735\tabularnewline
\hline 
32 & 36 & $1.971\times10^{-5}$ & 0.1598\tabularnewline
\hline
\end{tabular}

6 bits, no subroutines:

\begin{tabular}{|c|c|c|c|}
\hline 
Termination condition & $G$ & $D$ & $K$\tabularnewline
\hline 
40 & 2 & $5.31\times10^{-4}$ & 0.0461\tabularnewline
\hline 
48 & 11.5 & $5.07\times10^{-5}$ & 0.0819\tabularnewline
\hline
\end{tabular}

Termination condition 75\% (that is, the termination condition is
24 out of 32 for 5 bits, 48 out of 64 for 6 bits), no subroutines:

\begin{tabular}{|c|c|c|c|}
\hline 
Bits & $G$ & $D$ & $K$\tabularnewline
\hline 
5 & 3 & $4.48\times10^{-4}$ & 0.0635\tabularnewline
\hline 
6 & 11.5 & $5.07\times10^{-5}$ & 0.0819\tabularnewline
\hline
\end{tabular}

\pagebreak{}Termination condition 87.5\%, no subroutines:

\begin{tabular}{|c|c|c|c|}
\hline 
Bits & $G$ & $D$ & $K$\tabularnewline
\hline 
4 & 4 & $3.24\times10^{-4}$ & 0.072\tabularnewline
\hline 
5 & 12 & $3.75\times10^{-5}$ & 0.0735\tabularnewline
\hline
\end{tabular}

Termination condition 100\%, no subroutines:

\begin{tabular}{|c|c|c|c|}
\hline 
Bits & $G$ & $D$ & $K$\tabularnewline
\hline 
4 & 5 & $2.1\times10^{-4}$ & 0.0725\tabularnewline
\hline 
5 & 36 & $1.971\times10^{-5}$ & 0.1598\tabularnewline
\hline
\end{tabular}

4 bits, up to 4 subroutines:

\begin{tabular}{|c|c|c|c|}
\hline 
Termination condition & $G$ & $D$ & $K$\tabularnewline
\hline 
14 & 5 & $2.03\times10^{-4}$ & 0.0712\tabularnewline
\hline 
16 & 6 & $1.65\times10^{-4}$ & 0.0771\tabularnewline
\hline
\end{tabular}

5 bits, up to 4 subroutines:

\begin{tabular}{|c|c|c|c|}
\hline 
Termination condition & $G$ & $D$ & $K$\tabularnewline
\hline 
24 & 4 & $3.2\times10^{-4}$ & 0.0716\tabularnewline
\hline 
28 & 16.5 & $2.5\times10^{-5}$ & 0.0825\tabularnewline
\hline 
32 & 32 & $1.7\times10^{-5}$ & 0.1319\tabularnewline
\hline
\end{tabular}

6 bits, up to 4 subroutines:

\begin{tabular}{|c|c|c|c|}
\hline 
Termination condition & $G$ & $D$ & $K$\tabularnewline
\hline 
40 & 4 & $3.95\times10^{-4}$ & 0.0795\tabularnewline
\hline 
48 & 13.5 & $4.17\times10^{-5}$ & 0.0871\tabularnewline
\hline
\end{tabular}

Termination condition 75\%, up to 4 subroutines:

\begin{tabular}{|c|c|c|c|}
\hline 
Bits & $G$ & $D$ & $K$\tabularnewline
\hline 
5 & 4 & $3.2\times10^{-4}$ & 0.0716\tabularnewline
\hline 
6 & 13.5 & $4.167\times10^{-5}$ & 0.0871\tabularnewline
\hline
\end{tabular}

Termination condition 87.5\% up to 4 subroutines:

\begin{tabular}{|c|c|c|c|}
\hline 
Bits & $G$ & $D$ & $K$\tabularnewline
\hline 
4 & 5 & $2.03\times10^{-4}$ & 0.0712\tabularnewline
\hline 
5 & 16.5 & $2.5\times10^{-5}$ & 0.0825\tabularnewline
\hline
\end{tabular}

Termination condition 100\% up to 4 subroutines:

\begin{tabular}{|c|c|c|c|}
\hline 
Bits & $G$ & $D$ & $K$\tabularnewline
\hline 
4 & 6 & $1.65\times10^{-4}$ & 0.0771\tabularnewline
\hline 
5 & 32 & $1.7\times10^{-5}$ & 0.1319\tabularnewline
\hline
\end{tabular}

\subsection*{Statement set 2}

Statement set 2 contained the following node types: \texttt{And},
\texttt{Or}, \texttt{Nand}, \texttt{Nor}, and \texttt{Not}, plus the
constant nodes. The results were:

4 bits, no subroutines:

\begin{tabular}{|c|c|c|c|}
\hline 
Termination condition & $G$ & $D$ & $K$\tabularnewline
\hline 
10 & 7 & $3.67\times10^{-5}$ & 0.0424\tabularnewline
\hline 
11 & 53 & $1.512\times10^{-7}$ & 0.0206\tabularnewline
\hline
\end{tabular}

4 bits, up to 4 subroutines:

\begin{tabular}{|c|c|c|c|}
\hline 
Termination condition & $G$ & $D$ & $K$\tabularnewline
\hline 
10 & 7 & $2.7\times10^{-5}$ & 0.0364\tabularnewline
\hline 
11 & 66.5 & $1.09\times10^{-7}$ & 0.022\tabularnewline
\hline
\end{tabular}

\subsection*{Statement set 3}

Statement set 3 contained the following node types: \texttt{And},
\texttt{Or}, and \texttt{Not}, plus the constant nodes. The results
were:

4 bits, no subroutines:

\begin{tabular}{|c|c|c|c|}
\hline 
Termination condition & $G$ & $D$ & $K$\tabularnewline
\hline 
10 & 5 & $6.63\times10^{-5}$ & 0.0407\tabularnewline
\hline 
11 & 34 & $2.6\times10^{-7}$ & 0.01734\tabularnewline
\hline
\end{tabular}

4 bits, up to 4 subroutines:

\begin{tabular}{|c|c|c|c|}
\hline 
Termination condition & $G$ & $D$ & $K$\tabularnewline
\hline 
10 & 6 & $5.6\times10^{-5}$ & 0.0449\tabularnewline
\hline 
11 & 30 & $2.07\times10^{-7}$ & 0.01364\tabularnewline
\hline
\end{tabular}

\subsection*{Statement set 4}

Statement set 4 contained the following node types: \texttt{Plus},
\texttt{Minus}, \texttt{Times}, \texttt{Divide}, and \texttt{Negate}.
The constant nodes could take any integral value from $-3$ to 3.
Note that these operations treat variables as integers, not just as
booleans. In particular, \texttt{Divide} could throw a runtime exception
if the second operand evaluated to 0. If this occurred, the program
being executed was terminated and regarded as having gotten the wrong
answer for that set of inputs.

The results were:

4 bits, up to 4 subroutines:

\begin{tabular}{|c|c|c|c|}
\hline 
Termination condition & $G$ & $D$ & $K$\tabularnewline
\hline 
10 & 4 & $2.94\times10^{-4}$ & 0.0686\tabularnewline
\hline 
11 & 7 & $8.15\times10^{-5}$ & 0.0632\tabularnewline
\hline 
12 & 89.5 & $4.79\times10^{-7}$ & 0.062\tabularnewline
\hline
\end{tabular}

\subsection*{Statement set 5}

Statement set 5 contained the following node types: \texttt{Plus},
\texttt{Minus}, \texttt{Times}, \texttt{Divide}, \texttt{Negate},
\texttt{And}, \texttt{Or}, and \texttt{Not}, plus the constant nodes
from $-3$ to 3.. The results were:

4 bits, no subroutines:

\begin{tabular}{|c|c|c|c|}
\hline 
Termination condition & $G$ & $D$ & $K$\tabularnewline
\hline 
11 & 6 & $1.36\times10^{-4}$ & 0.07\tabularnewline
\hline 
12 & 9 & $4.43\times10^{-5}$ & 0.0599\tabularnewline
\hline
\end{tabular}

Statement sets 2, 3, 4, and 5 seemed to scale better than $\frac{1}{\sqrt{D}}$,
rather than worse. But a look at the evolved programs revealed that
in each case, their length distribution departed from the expected
length distribution (the distribution that was used to generate the
programs). This is not surprising, since the expected length distribution
of generated programs was created to match the distribution of working
programs shown by statement set 1. Also, statement sets 2 and 4 departed
from the expected distribution for the number of subroutines.

I created a new length distribution for statement set 5. I re-ran
both the evolutions and the density measurements with this new length
distribution, with the following results:

\begin{tabular}{|c|c|c|c|}
\hline 
Termination condition & $G$ & $D$ & $K$\tabularnewline
\hline 
11 & 5 & $2.19\times10^{-4}$ & 0.074\tabularnewline
\hline 
12 & 8 & $6.77\times10^{-5}$ & 0.0658\tabularnewline
\hline 
13 & 32 & $9.33\times10^{-6}$ & 0.0978\tabularnewline
\hline 
14 & 44 & $6.03\times10^{-6}$ & 0.108\tabularnewline
\hline 
15 & 209 & $1.49\times10^{-6}$ & 0.255\tabularnewline
\hline
\end{tabular}

It must be noted, however, that at higher termination conditions,
the length profile of the working programs still did not match the
new length profile of the generated programs, so the validity of this
data is suspect.

\section{PARALLEL SYSTEMS}

The first two systems (sorting and parity) are classic computer science
problems. In each case, the output is a very complex function of the
input. (Actually, in the sorting case, correct output is always the
same, but the transformation from input to output is very complex.)
For the third system, I chose a very smooth function - an $n$-dimensional
Gaussian curve, centered at the origin in the $n$-dimensional cube
which extended over the interval $[-1,1)$ in each dimension. (The
exclusion of 1.0 was an artifact of the means of generating random
real numbers; it does not seem possible for it to affect the results.)

{}``Programs'' were really data, represented as an $n$-vector lying
within the $n$-dimensional cube. Obviously, programs were of fixed
length. Unlike the previous system, this system (and all subsequent
ones) mutated at most one element of the vector.

The first two systems were \emph{program-like} - the programs looked
like statements to be executed. In contrast, the third system was
\emph{data-like} - programs looked like coordinates at which a function
was to be evaluated.

A program's fitness function was $e^{-r^{2}}$, where $r$ was the
Euclidean distance from the program's vector to the origin. If the
fitness function was equal to or greater than the termination value,
the program was considered to be fully working. (Perfection - a fitness
function of $1.0$ - was not realistically achievable for this system.)

Unfortunately, sometimes the density became so low that it could not
be measured by the standard Monte Carlo method that was used on previous
systems (at least not within a reasonable amount of CPU time). However,
the density can be calculated. If the termination threshold is t,
then all points in the $n$-dimensional space that lie inside the
sphere with radius $r=\sqrt{-ln(t)}$ meet the termination condition.
For $n$ dimensions, the volume $V$ of the sphere is given by $V=\frac{(2\pi)^{n/2}r^{n}}{2\times4\times\ldots\times n}$
for even $n$, and $V=2\frac{(2\pi)^{(n-1)/2}r^{n}}{1\times3\times\ldots\times n}$
for odd $n$. The volume of the entire space is $2^{n}$, since it
extends from $-1$ to 1 in all $n$ dimensions. The density of working
programs is then $D=\frac{V}{2^{n}}$. These results use the calculated
density exclusively.

Here are the results for this system:

Two dimensions:

\begin{tabular}{|c|c|c|c|}
\hline 
Termination condition & $G$ & $D$ & $K$\tabularnewline
\hline 
0.99 & 26 & $7.89\times10^{-3}$ & 2.31\tabularnewline
\hline 
0.999 & 320 & $7.86\times10^{-4}$ & 8.97\tabularnewline
\hline 
0.9999 & 1036 & $7.85\times10^{-5}$ & 9.18\tabularnewline
\hline 
0.99999 & 3702 & $7.85\times10^{-6}$ & 10.37\tabularnewline
\hline 
0.999999 & 14812.5 & $7.85\times10^{-7}$ & 13.13\tabularnewline
\hline 
0.9999999 & 40284 & $7.85\times10^{-8}$ & 11.29\tabularnewline
\hline
\end{tabular}

Four dimensions:

\begin{tabular}{|c|c|c|c|}
\hline 
Termination condition & $G$ & $D$ & $K$\tabularnewline
\hline 
0.99 & 328.5 & $3.12\times10^{-5}$ & 1.834\tabularnewline
\hline 
0.999 & 1262.5 & $3.09\times10^{-7}$ & 0.701\tabularnewline
\hline 
0.9999 & 3968 & $3.08\times10^{-9}$ & 0.22\tabularnewline
\hline 
0.99999 & 16427 & $3.08\times10^{-11}$ & 0.0912\tabularnewline
\hline 
0.999999 & 47237.5 & $3.08\times10^{-13}$ & 0.0262\tabularnewline
\hline 
0.9999999 & 140100 & $3.08\times10^{-15}$ & 0.00778\tabularnewline
\hline
\end{tabular}

Six dimensions:

\begin{tabular}{|c|c|c|c|}
\hline 
Termination condition & $G$ & $D$ & $K$\tabularnewline
\hline 
0.8 & 26 & $8.97\times10^{-4}$ & 0.779\tabularnewline
\hline 
0.9 & 120.5 & $9.44\times10^{-5}$ & 1.171\tabularnewline
\hline 
0.99 & 825 & $8.2\times10^{-8}$ & 0.236\tabularnewline
\hline 
0.999 & 2516 & $8.09\times10^{-11}$ & 0.0226\tabularnewline
\hline 
0.9999 & 8562 & $8.08\times10^{-14}$ & $2.43\times10^{-3}$\tabularnewline
\hline 
0.99999 & 29224.5 & $8.07\times10^{-17}$ & $2.63\times10^{-4}$\tabularnewline
\hline 
0.999999 & 86167.5 & $8.07\times10^{-20}$ & $2.45\times10^{-5}$\tabularnewline
\hline 
0.9999999 & 276830.5 & $8.07\times10^{-23}$ & $2.49\times10^{-6}$\tabularnewline
\hline
\end{tabular}

\pagebreak{}Eight dimensions:

\begin{tabular}{|c|c|c|c|}
\hline 
Termination condition & $G$ & $D$ & $K$\tabularnewline
\hline 
0.6 & 7.5 & $1.08\times10^{-3}$ & 0.246\tabularnewline
\hline 
0.7 & 50.5 & $2.57\times10^{-4}$ & 0.809\tabularnewline
\hline 
0.8 & 117.5 & $3.93\times10^{-5}$ & 0.737\tabularnewline
\hline 
0.9 & 264.5 & $1.954\times10^{-6}$ & 0.37\tabularnewline
\hline 
0.99 & 1151.5 & $1.618\times10^{-10}$ & 0.01465\tabularnewline
\hline 
0.999 & 4343.5 & $1.589\times10^{-14}$ & $5.47\times10^{-4}$\tabularnewline
\hline 
0.9999 & 13637.5 & $1.586\times10^{-18}$ & $1.717\times10^{-5}$\tabularnewline
\hline 
0.99999 & 42908 & $1.586\times10^{-22}$ & $5.4\times10^{-7}$\tabularnewline
\hline 
0.999999 & 147670.5 & $1.586\times10^{-26}$ & $1.859\times10^{-8}$\tabularnewline
\hline 
0.9999999 & 440362.5 & $1.585\times10^{-30}$ & $5.54\times10^{-10}$\tabularnewline
\hline
\end{tabular}

Ten dimensions:

\begin{tabular}{|c|c|c|c|}
\hline 
Termination condition & $G$ & $D$ & $K$\tabularnewline
\hline 
0.5 & 5 & $3.98\times10^{-4}$ & 0.0998\tabularnewline
\hline 
0.6 & 66.5 & $8.66\times10^{-5}$ & 0.619\tabularnewline
\hline 
0.7 & 147 & $1.438\times10^{-5}$ & 0.557\tabularnewline
\hline 
0.8 & 247 & $1.379\times10^{-6}$ & 0.29\tabularnewline
\hline 
0.9 & 449 & $3.23\times10^{-8}$ & 0.0807\tabularnewline
\hline 
0.99 & 1983 & $2.55\times10^{-13}$ & $1.002\times10^{-3}$\tabularnewline
\hline 
0.999 & 6779 & $2.5\times10^{-18}$ & $1.071\times10^{-5}$\tabularnewline
\hline 
0.9999 & 19896 & $2.49\times10^{-23}$ & $9.93\times10^{-8}$\tabularnewline
\hline 
0.99999 & 60303 & $2.49\times10^{-28}$ & $9.52\times10^{-10}$\tabularnewline
\hline 
0.999999 & 195696 & $2.49\times10^{-33}$ & $9.77\times10^{-12}$\tabularnewline
\hline 
0.9999999 & 631803.5 & $2.49\times10^{-38}$ & $9.97\times10^{-14}$\tabularnewline
\hline
\end{tabular}

Obviously, this system did \emph{not} demonstrate the {}``slightly
worse than $\frac{1}{\sqrt{D}}$'' behavior that the first two systems
showed!

Other than smoothness, the Gaussian system has another difference
from the first two systems: if all but one of the variables are held
constant, the result is a one-dimensional Gaussian curve in the remaining
variable. Further, the one-dimensional Gaussian curve and the multi-dimensional
Gaussian curve are centered at the same value of the non-constant
variable. This means that the Gaussian system can optimize each variable
independently. In general, the first two systems could not do this.

The Gaussian system can therefore be considered a \emph{parallel}
system, in that it is conducting $n$ essentially independent evolutions
in parallel, with the results all multiplied together into one fitness
function.

To explore such systems further, I built a second parallel system.
The {}``program'' for this system consisted of $n$ variables, each
of which could range from 1 to $p$. The fitness function was the
number of variables that had value $p$ (hence this system may be
called the {}``highest'' system). No termination condition was used
for this system; an $n$-dimensional program had to have a fitness
function equal to $n$ to be considered working. The density is therefore
$D=\frac{1}{p^{n}}$.

Here are the results for this system:

Two dimensions:

\begin{tabular}{|c|c|c|c|}
\hline 
$p$ & $G$ & $D$ & $K$\tabularnewline
\hline 
50 & 459 & $4.0\times10^{-4}$ & 9.18\tabularnewline
\hline 
100 & 1069.5 & $1.0\times10^{-4}$ & 10.7\tabularnewline
\hline 
200 & 2390.5 & $2.5\times10^{-5}$ & 11.95\tabularnewline
\hline 
500 & 5444.5 & $4.0\times10^{-6}$ & 10.89\tabularnewline
\hline 
1000 & 10124 & $1.0\times10^{-6}$ & 10.12\tabularnewline
\hline 
2000 & 22724.5 & $2.5\times10^{-7}$ & 11.36\tabularnewline
\hline
\end{tabular}

Four dimensions:

\begin{tabular}{|c|c|c|c|}
\hline 
$p$ & $G$ & $D$ & $K$\tabularnewline
\hline 
50 & 1435 & $1.6\times10^{-7}$ & 0.574\tabularnewline
\hline 
100 & 3121.5 & $1.0\times10^{-8}$ & 0.312\tabularnewline
\hline 
200 & 7161 & $6.25\times10^{-10}$ & 0.179\tabularnewline
\hline 
500 & 17504.5 & $1.6\times10^{-11}$ & 0.07\tabularnewline
\hline 
1000 & 37920.5 & $1.0\times10^{-12}$ & 0.0379\tabularnewline
\hline 
2000 & 76007.5 & $6.25\times10^{-14}$ & 0.019\tabularnewline
\hline
\end{tabular}

Six dimensions:

\begin{tabular}{|c|c|c|c|}
\hline 
$p$ & $G$ & $D$ & $K$\tabularnewline
\hline 
50 & 2794.5 & $6.4\times10^{-11}$ & 0.0224\tabularnewline
\hline 
100 & 5805.5 & $1.0\times10^{-12}$ & $5.81\times10^{-3}$\tabularnewline
\hline 
200 & 12204.5 & $1.562\times10^{-14}$ & $1.524\times10^{-3}$\tabularnewline
\hline 
500 & 32522 & $6.4\times10^{-17}$ & $2.6\times10^{-4}$\tabularnewline
\hline 
1000 & 62569.5 & $1.0\times10^{-18}$ & $6.26\times10^{-5}$\tabularnewline
\hline 
2000 & 136637.5 & $1.562\times10^{-20}$ & $1.707\times10^{-5}$\tabularnewline
\hline
\end{tabular}

Eight dimensions:

\begin{tabular}{|c|c|c|c|}
\hline 
$p$ & $G$ & $D$ & $K$\tabularnewline
\hline 
50 & 3899.5 & $2.56\times10^{-14}$ & $6.24\times10^{-4}$\tabularnewline
\hline 
100 & 8705.5 & $1.0\times10^{-16}$ & $8.71\times10^{-5}$\tabularnewline
\hline 
200 & 17692 & $3.91\times10^{-19}$ & $1.106\times10^{-5}$\tabularnewline
\hline 
500 & 52573.5 & $2.56\times10^{-22}$ & $8.41\times10^{-7}$\tabularnewline
\hline 
1000 & 93348.5 & $1.0\times10^{-24}$ & $9.33\times10^{-8}$\tabularnewline
\hline 
2000 & 204464.5 & $3.91\times10^{-27}$ & $1.278\times10^{-8}$\tabularnewline
\hline
\end{tabular}

Ten dimensions:

\begin{tabular}{|c|c|c|c|}
\hline 
$p$ & $G$ & $D$ & $K$\tabularnewline
\hline 
50 & 5511 & $1.024\times10^{-17}$ & $1.764\times10^{-5}$\tabularnewline
\hline 
100 & 13178 & $1.0\times10^{-20}$ & $1.318\times10^{-6}$\tabularnewline
\hline 
200 & 26675.5 & $9.77\times10^{-24}$ & $8.34\times10^{-8}$\tabularnewline
\hline 
500 & 66886 & $1.024\times10^{-27}$ & $2.14\times10^{-9}$\tabularnewline
\hline 
1000 & 143047 & $1.0\times10^{-30}$ & $1.43\times10^{-10}$\tabularnewline
\hline 
2000 & 268494 & $9.77\times10^{-34}$ & $8.39\times10^{-12}$\tabularnewline
\hline
\end{tabular}

Once again, this system did not demonstrate the {}``slightly worse
than $\frac{1}{\sqrt{D}}$'' behavior that the first two systems
showed. But there is more to see here. Looking just at the median
number of generations to evolve a working program, we see that the
ratio between the number of generations for $p=2000$ and the number
of generations for $p=50$ stayed remarkably consistent as the number
of dimensions changed. It looks like $G$ might be separable into
a component that depends on $p$ and a component that depends on the
number of dimensions ($n$), that is, $G=f_{1}(p)\times f_{2}(n)$,
for some $f_{1}$ and $f_{2}$.

Empirically, $f_{2}(n)\approxeq\frac{1}{n'\times\ln(n')}$, where
$n'=n+0.6$.

If all the variation for the number of dimensions is in $f_{2}$,
then $f_{1}$ must be the same for all values of $n$. In particular,
it must be the same for $n=2$. So it seems reasonable for $f_{1}$
to depend only on the density for $n=2$. Is $f_{1}$ the familiar
{}``slightly worse than $\frac{1}{\sqrt{D}}$'' behavior?

Let $K'(\delta)=\frac{G\times\sqrt{D_{2}}}{(n+\delta)\ln(n+\delta)}$,
where $D_{2}$ is the density for $n=2$. That is, $\frac{1}{\sqrt{D_{2}}}$
is a candidate for $f_{2}$. For this system, we get the following
results:

Two dimensions:

\begin{tabular}{|c|c|c|c|}
\hline 
$p$ & $G$ & $D_{2}$ & $K'(0.6)$\tabularnewline
\hline 
50 & 459 & $4.0\times10^{-4}$ & 3.7\tabularnewline
\hline 
100 & 1069.5 & $1.0\times10^{-4}$ & 4.3\tabularnewline
\hline 
200 & 2390.5 & $2.5\times10^{-5}$ & 4.81\tabularnewline
\hline 
500 & 5444.5 & $4.0\times10^{-6}$ & 4.38\tabularnewline
\hline 
1000 & 10124 & $1.0\times10^{-6}$ & 4.08\tabularnewline
\hline 
2000 & 22724.5 & $2.5\times10^{-7}$ & 4.57\tabularnewline
\hline
\end{tabular}

Four dimensions:

\begin{tabular}{|c|c|c|c|}
\hline 
$p$ & $G$ & $D_{2}$ & $K'(0.6)$\tabularnewline
\hline 
50 & 1435 & $4.0\times10^{-4}$ & 4.09\tabularnewline
\hline 
100 & 3121.5 & $1.0\times10^{-4}$ & 4.45\tabularnewline
\hline 
200 & 7161 & $2.5\times10^{-5}$ & 5.1\tabularnewline
\hline 
500 & 17504.5 & $4.0\times10^{-6}$ & 4.99\tabularnewline
\hline 
1000 & 37920.5 & $1.0\times10^{-6}$ & 5.4\tabularnewline
\hline 
2000 & 76007.5 & $2.5\times10^{-7}$ & 5.41\tabularnewline
\hline
\end{tabular}

Six dimensions:

\begin{tabular}{|c|c|c|c|}
\hline 
$p$ & $G$ & $D_{2}$ & $K'(0.6)$\tabularnewline
\hline 
50 & 2794.5 & $4.0\times10^{-4}$ & 4.49\tabularnewline
\hline 
100 & 5805.5 & $1.0\times10^{-4}$ & 4.66\tabularnewline
\hline 
200 & 12204.5 & $2.5\times10^{-5}$ & 4.9\tabularnewline
\hline 
500 & 32522 & $4.0\times10^{-6}$ & 5.22\tabularnewline
\hline 
1000 & 62569.5 & $1.0\times10^{-6}$ & 5.02\tabularnewline
\hline 
2000 & 136637.5 & $2.5\times10^{-7}$ & 5.49\tabularnewline
\hline
\end{tabular}

\pagebreak{}Eight dimensions:

\begin{tabular}{|c|c|c|c|}
\hline 
$p$ & $G$ & $D_{2}$ & $K'(0.6)$\tabularnewline
\hline 
50 & 3899.5 & $4.0\times10^{-4}$ & 4.21\tabularnewline
\hline 
100 & 8705.5 & $1.0\times10^{-4}$ & 4.7\tabularnewline
\hline 
200 & 17692 & $2.5\times10^{-5}$ & 4.78\tabularnewline
\hline 
500 & 52573.5 & $4.0\times10^{-6}$ & 5.68\tabularnewline
\hline 
1000 & 93348.5 & $1.0\times10^{-6}$ & 5.04\tabularnewline
\hline 
2000 & 204464.5 & $2.5\times10^{-7}$ & 5.52\tabularnewline
\hline
\end{tabular}

Ten dimensions:

\begin{tabular}{|c|c|c|c|}
\hline 
$p$ & $G$ & $D_{2}$ & $K'(0.6)$\tabularnewline
\hline 
50 & 5511 & $4.0\times10^{-4}$ & 4.4\tabularnewline
\hline 
100 & 13178 & $1.0\times10^{-4}$ & 5.27\tabularnewline
\hline 
200 & 26675.5 & $2.5\times10^{-5}$ & 5.33\tabularnewline
\hline 
500 & 66886 & $4.0\times10^{-6}$ & 5.35\tabularnewline
\hline 
1000 & 143047 & $1.0\times10^{-6}$ & 5.72\tabularnewline
\hline 
2000 & 268494 & $2.5\times10^{-7}$ & 5.36\tabularnewline
\hline
\end{tabular}

Turning back to the Gaussian system, we find that it has a good fit
with $f_{2}(n)\approxeq\frac{1}{n''\times\ln(n'')}$, where $n''=n+0.05$.
We get the following results for the Gaussian system:

Two dimensions:

\begin{tabular}{|c|c|c|c|}
\hline 
Termination condition & $G$ & $D_{2}$ & $K'(0.05)$\tabularnewline
\hline 
0.99 & 26 & $7.89\times10^{-3}$ & 1.57\tabularnewline
\hline 
0.999 & 320 & $7.86\times10^{-4}$ & 6.1\tabularnewline
\hline 
0.9999 & 1036 & $7.85\times10^{-5}$ & 6.24\tabularnewline
\hline 
0.99999 & 3702 & $7.85\times10^{-6}$ & 7.05\tabularnewline
\hline 
0.999999 & 14812.5 & $7.85\times10^{-7}$ & 8.92\tabularnewline
\hline 
0.9999999 & 40284 & $7.85\times10^{-8}$ & 7.67\tabularnewline
\hline
\end{tabular}

Four dimensions:

\begin{tabular}{|c|c|c|c|}
\hline 
Termination condition & $G$ & $D_{2}$ & $K'(0.05)$\tabularnewline
\hline 
0.99 & 328.5 & $7.89\times10^{-3}$ & 5.15\tabularnewline
\hline 
0.999 & 1262.5 & $7.86\times10^{-4}$ & 6.25\tabularnewline
\hline 
0.9999 & 3968 & $7.85\times10^{-5}$ & 6.21\tabularnewline
\hline 
0.99999 & 16427 & $7.85\times10^{-6}$ & 8.13\tabularnewline
\hline 
0.999999 & 47237.5 & $7.85\times10^{-7}$ & 7.39\tabularnewline
\hline 
0.9999999 & 140100 & $7.85\times10^{-8}$ & 6.93\tabularnewline
\hline
\end{tabular}

\pagebreak{}Six dimensions:

\begin{tabular}{|c|c|c|c|}
\hline 
Termination condition & $G$ & $D_{2}$ & $K'(0.05)$\tabularnewline
\hline 
0.8 & 26 & 0.1753 & 0.999\tabularnewline
\hline 
0.9 & 120.5 & 0.0827 & 3.18\tabularnewline
\hline 
0.99 & 825 & $7.89\times10^{-3}$ & 6.73\tabularnewline
\hline 
0.999 & 2516 & $7.86\times10^{-4}$ & 6.48\tabularnewline
\hline 
0.9999 & 8562 & $7.85\times10^{-5}$ & 6.97\tabularnewline
\hline 
0.99999 & 29224.5 & $7.85\times10^{-6}$ & 7.52\tabularnewline
\hline 
0.999999 & 86167.5 & $7.85\times10^{-7}$ & 7.01\tabularnewline
\hline 
0.9999999 & 276830.5 & $7.85\times10^{-8}$ & 7.12\tabularnewline
\hline
\end{tabular}

Eight dimensions:

\begin{tabular}{|c|c|c|c|}
\hline 
Termination condition & $G$ & $D_{2}$ & $K'(0.05)$\tabularnewline
\hline 
0.6 & 7.5 & 0.401 & 0.283\tabularnewline
\hline 
0.7 & 50.5 & 0.28 & 1.592\tabularnewline
\hline 
0.8 & 117.5 & 0.1753 & 2.93\tabularnewline
\hline 
0.9 & 264.5 & 0.0827 & 4.53\tabularnewline
\hline 
0.99 & 1151.5 & $7.89\times10^{-3}$ & 6.09\tabularnewline
\hline 
0.999 & 4343.5 & $7.86\times10^{-4}$ & 7.25\tabularnewline
\hline 
0.9999 & 13637.5 & $7.85\times10^{-5}$ & 7.2\tabularnewline
\hline 
0.99999 & 42908 & $7.85\times10^{-6}$ & 7.16\tabularnewline
\hline 
0.999999 & 147670.5 & $7.85\times10^{-7}$ & 7.79\tabularnewline
\hline 
0.9999999 & 440362.5 & $7.85\times10^{-8}$ & 7.35\tabularnewline
\hline
\end{tabular}

Ten dimensions:

\begin{tabular}{|c|c|c|c|}
\hline 
Termination condition & $G$ & $D_{2}$ & $K'(0.05)$\tabularnewline
\hline 
0.5 & 5 & 0.544 & 0.1591\tabularnewline
\hline 
0.6 & 66.5 & 0.401 & 1.816\tabularnewline
\hline 
0.7 & 147 & 0.28 & 3.35\tabularnewline
\hline 
0.8 & 247 & 0.1753 & 4.46\tabularnewline
\hline 
0.9 & 449 & 0.0827 & 5.57\tabularnewline
\hline 
0.99 & 1983 & $7.89\times10^{-3}$ & 7.6\tabularnewline
\hline 
0.999 & 6779 & $7.86\times10^{-4}$ & 8.19\tabularnewline
\hline 
0.9999 & 19896 & $7.85\times10^{-5}$ & 7.6\tabularnewline
\hline 
0.99999 & 60303 & $7.85\times10^{-6}$ & 7.29\tabularnewline
\hline 
0.999999 & 195696 & $7.85\times10^{-7}$ & 7.48\tabularnewline
\hline 
0.9999999 & 631803.5 & $7.85\times10^{-8}$ & 7.63\tabularnewline
\hline
\end{tabular}

These results are similar to those of the {}``highest'' system,
with the $K'$ values relatively flat for the same termination condition,
independent of dimension. Also, as the problem gets harder, $K'$
exhibits the {}``slightly worse than $\frac{1}{\sqrt{D}}$'' behavior
- though it seems steeper than the first two systems for low termination
conditions and flatter for high termination conditions.

At higher termination conditions, the usual {}``slightly worse than
$\frac{1}{\sqrt{D}}$'' behavior is clearer if we use the density
from $n=1$ (one dimension) rather than from $n=2$. But the argument
for why we can use the density from $n=2$ may not be valid when applied
to $n=1$, since a system with only one dimension can only evolve
by mutation. While I suspect that this makes no difference for the
density data, it seems safer to use the data from $n=2$.

\section{ANTI-PARALLEL (TWISTED) SYSTEMS}

A parallel system is one where the fitness function can be optimized
for each dimension independently. In contrast, a system where the
fitness function \emph{must} be optimized for all dimensions simultaneously
may be called \emph{anti-parallel}. (Attempting to optimize just one
variable gives the wrong answer for that variable.) I built three
such systems.

The {}``binary'' system had $n$ dimensions and $b$ bits. Each
dimension had an integer variable, with a range from 1 to $2^{b}-1$.
(In practice, it was implemented with the range from 0 to $2^{b}-2$,
to be able to use a 0-based lookup table for the fitness.) The fitness
value of each dimension was the value of that dimension's variable,
reduced to just the least set bit. For example, with $b=4$, the values
were $\left\{ 1,2,1,4,1,2,1,8,1,2,1,4,1,2,1\right\} $. The value
of the fitness function was the sum of the fitness values for each
dimension. This system is a parallel system.

To make a parallel system into an anti-parallel system, we perform
a rotation of axes. Given that in the binary system, the variables
must be integers in the correct range, I could not use the obvious
transformation, which is $u=\frac{(x+y)}{\sqrt{2}}$, $v=\frac{(x-y)}{\sqrt{2}}$.
Instead, I used $u=\frac{(x+y)}{2}$ (integer division, that is, discarding
any remainder), and $v=\left|x-y\right|$. (This is another place
where using a range that starts at 0 helped the implementation.) I
paired the dimensions to do this; this required that $n$ be even.
This means that this system was only anti-parallel for pairs of dimensions.
As the investigation of this system did not proceed beyond $n=2$,
this was not an issue.

This system could not evolve. If it didn't have a working program
in the first generation, it almost always got permanently stuck. With
20 programs, about $\frac{1}{3}$ of the time it could not evolve
$n=2$, $b=3$ with a termination value of 8, even though there were
only 49 programs possible with those values, and two of them met the
termination criterion! ({}``Permanently stuck'' is impossible to
prove. However, it got stuck with the best program having a fitness
value of 6 for 100,000,000 generations; at that point the evolution
was terminated. That was as close to {}``permanent'' as I had patience
for.)

It is easy to see why the twisted binary system got stuck. With two
dimensions and three bits, to get a fitness function of 6, $(x,y)$
must be one of $(0,3)$, $(3,0)$, $(3,4)$, $(4,3)$, $(1,6)$, or
$(6,1)$. For a fitness function of 8, $(x,y)$ must be $(2,5)$ or
$(5,2)$. (Note that these are 0-based $x$ and $y$, not 1-based.)
If the entire population reaches a fitness function of 6, the evolution
is stuck. It must change both $x$ and $y$ to reach a fitness function
of 8, it cannot get either $x$ or $y$ to the right value by any
combination of parents, it can only mutate $x$ or $y$ (but not both)
in one generation, and the result of mutating either $x$ or $y$
is a less-fit offspring. Also, a fitness value of 7 is not possible
in this system, so there is no way to reach 8 by taking two steps.

The only possible way forward would be for there to be at least seven
programs with a mutation in the previous generation (probability $10^{-2}$
for each one, so $10^{-14}$ for all seven, but it could be any seven
out of the population of 20, which gives us 77520 ways that it could
happen, for a total probability of $7.752\times10^{-10}$). Then all
seven mutated programs would have to be chosen for the competition
for a parent of a program in the next generation (one out of 77520,
but there are 40 such competitions, so the probability is $\frac{40}{77520}$
- though this is not quite exact). Then the mutation would have to
be passed on to the next generation (probability 0.5). Then there
would have to be \emph{another} mutation in the child program (probability
0.01). Finally, the mutations would have to give rise to the right
values so that the resulting fitness function was 8 (probability $\frac{2}{7}$
for the first mutation, and $\frac{1}{7}$ for the second). This combination
of events is immensely unlikely (total probability $8.16\times10^{-17}$per
generation).

The second anti-parallel system I build was based on the {}``linear''
system. This system had $n$ dimensions. Each dimension had a real
variable, in the range $\left[-1.0,1.0\right)$. The fitness function
was 1, minus the sum of the absolute values of the variable for each
dimension. This {}``linear'' system was a parallel system.

To convert the linear system into an anti-parallel system, I rotated
it through 45 degrees in each of the Euler angles. I then scaled the
rotated variables by different amounts: the first rotated variable
was scaled by $1.0$, the second by $1.5$, the third by $1.5^{2}$,
and so on. The fitness function was 1, minus the sum of the absolute
values of the rotated and scaled variables. This created, essentially,
a diagonal {}``ridge'', with the fitness function falling away more
steeply in other directions. This {}``twisted linear'' system was
anti-parallel.

It may be easier to see why this system is anti-parallel in the two-dimensional
case. Holding one variable constant defines a line; optimizing the
other variable means finding the highest point on the line, which
is where the line intersects the ridge. But because the line is parallel
to one of the axes and the ridge is diagonal, this gives a value for
the optimized variable that is different from the coordinate of the
peak (the highest point on the ridge).

This system could barely evolve at all. Starting at two dimensions
with termination value $0.7$, it occasionally took half a million
generations to evolve a solution, even though the solution has a density
that was greater than 1\%. At termination value $0.8$, it once only
made it to fitness = $0.67$ in 112 million generations. The evolution
was terminated at that point. This was slightly better than the twisted
binary system, but it still essentially could not evolve anything
more than the most trivial problems. This happens because of the shape
of the fitness function. The absolute values cause a discontinuous
first derivative at the ridge. From a point on the ridge, moving in
any direction parallel to an axis (that is, changing any one variable)
reduces the fitness function. Also, from a point near the ridge, the
only way to improve the fitness function is to move closer to the
ridge.

The third anti-parallel system was created by applying the rotations
and scaling of the twisted linear system to the Gaussian system. The
fitness function was $e^{-R^{2}}$, where $R$ was the Euclidean length
of the vector composed of the rotated and scaled variables.

Like the Gaussian system, the density of this system was easy to calculate.
It was the same as the density of the Gaussian system, except for
the scaling. These results use the calculated density exclusively.

Here are the results for this system:

\pagebreak{}Two dimensions:

\begin{tabular}{|c|c|c|c|c|c|}
\hline 
Termination condition & $G$ & $D$ & $K$ & $K'(0.0)$ & $K'(1.0)$\tabularnewline
\hline 
0.95 & 2 & 0.0269 & 0.328 & 0.688 & 0.289\tabularnewline
\hline 
0.99 & 10 & $5.26\times10^{-3}$ & 0.725 & 2.28 & 0.961\tabularnewline
\hline
\end{tabular}

Four dimensions:

\begin{tabular}{|c|c|c|c|c|c|}
\hline 
Termination condition & $G$ & $D$ & $K$ & $K'(0.0)$ & $K'(1.0)$\tabularnewline
\hline 
0.5 & 2 & 0.01301 & 0.228 & 0.329 & 0.227\tabularnewline
\hline 
0.6 & 3 & $7.07\times10^{-3}$ & 0.252 & 0.457 & 0.315\tabularnewline
\hline 
0.7 & 47.5 & $3.44\times10^{-3}$ & 2.79 & 6.62 & 4.56\tabularnewline
\hline 
0.8 & 99 & $1.348\times10^{-4}$ & 3.64 & 12.27 & 8.46\tabularnewline
\hline 
0.9 & 299.5 & $3.01\times10^{-4}$ & 5.19 & 30.8 & 21.2\tabularnewline
\hline 
0.95 & 637 & $7.12\times10^{-5}$ & 5.38 & 54.7 & 37.7\tabularnewline
\hline 
0.99 & 1595 & $2.74\times10^{-6}$ & 2.64 & 91.1 & 62.8\tabularnewline
\hline
\end{tabular}

Six dimensions:

\begin{tabular}{|c|c|c|c|c|c|}
\hline 
Termination condition & $G$ & $D$ & $K$ & $K'(0.0)$ & $K'(1.0)$\tabularnewline
\hline 
0.5 & 614.5 & $6.14\times10^{-5}$ & 4.82 & 52.2 & 41.2\tabularnewline
\hline 
0.6 & 972 & $2.46\times10^{-5}$ & 4.82 & 76.4 & 60.3\tabularnewline
\hline 
0.7 & 1341.5 & $8.37\times10^{-6}$ & 3.88 & 96.4 & 76.1\tabularnewline
\hline 
0.8 & 1825.5 & $2.05\times10^{-6}$ & 2.61 & 116.7 & 92.1\tabularnewline
\hline 
0.9 & 4782.5 & $2.16\times10^{-7}$ & 2.22 & 253 & 200\tabularnewline
\hline 
0.95 & 7144.5 & $2.49\times10^{-8}$ & 1.127 & 316 & 250\tabularnewline
\hline 
0.99 & 23444.5 & $1.872\times10^{-10}$ & 0.321 & 690 & 545\tabularnewline
\hline
\end{tabular}

Eight dimensions:

\begin{tabular}{|c|c|c|c|c|c|}
\hline 
Termination condition & $G$ & $D$ & $K$ & $K'(0.0)$ & $K'(1.0)$\tabularnewline
\hline 
0.5 & 10868.5 & $4.29\times10^{-8}$ & 2.25 & 596 & 501\tabularnewline
\hline 
0.6 & 12296 & $1.267\times10^{-8}$ & 1.384 & 625 & 526\tabularnewline
\hline 
0.7 & 22758 & $3.01\times10^{-9}$ & 1.249 & 1057 & 889\tabularnewline
\hline 
0.8 & 39954.5 & $4.61\times10^{-10}$ & 0.858 & 1651 & 1389\tabularnewline
\hline 
0.9 & 70855.5 & $2.29\times10^{-11}$ & 0.339 & 2430 & 2040\tabularnewline
\hline 
0.95 & 183781 & $1.288\times10^{-12}$ & 0.208 & 5260 & 4420\tabularnewline
\hline 
0.99 & 580747.5 & $1.898\times10^{-15}$ & 0.0253 & 11050 & 9300\tabularnewline
\hline
\end{tabular}

Ten dimensions:

\begin{tabular}{|c|c|c|c|c|c|}
\hline 
Termination condition & $G$ & $D$ & $K$ & $K'(0.0)$ & $K'(1.0)$\tabularnewline
\hline 
0.5 & 215107.5 & $4.75\times10^{-12}$ & 0.469 & 8520 & 7440\tabularnewline
\hline 
0.6 & 319606.5 & $1.032\times10^{-12}$ & 0.325 & 11730 & 10240\tabularnewline
\hline 
0.7 & 468323 & $1.712\times10^{-13}$ & 0.1938 & 15720 & 13720\tabularnewline
\hline 
0.8 & 947594.5 & $1.641\times10^{-14}$ & 0.1214 & 28300 & 24700\tabularnewline
\hline 
0.9 & 2717644 & $3.85\times10^{-16}$ & 0.0533 & 67200 & 58700\tabularnewline
\hline 
0.95 & 4389708.5 & $1.053\times10^{-17}$ & 0.01425 & 90700 & 79200\tabularnewline
\hline 
0.99 & 14162811.5 & $3.04\times10^{-21}$ & $7.81\times10^{-4}$ & 194800 & 170000\tabularnewline
\hline
\end{tabular}

This system clearly scaled better than the first two systems. Even
without knowing the appropriate value of $\delta$ to use, we can
see that this system scaled worse than a parallel system. Finally,
it scaled much better than the twisted binary and twisted linear systems.
This is because the twisted Gaussian system has continuous derivatives.
The gradient is nonzero everywhere except at the peak. This means
that, unlike the twisted linear and twisted binary systems, the twisted
Gaussian system could always make progress by changing only one variable.

\section{CONCLUSIONS}

Genetic programming scales very well for data-like problems with continuous
first derivatives (except for the problem of getting stuck on a sub-peak).
But for program-like problems, genetic programming doesn't seem to
scale very well to larger, more difficult problems. As the size of
the solution space increases, the number of working programs also
increases, but more slowly. So the density of working programs decreases,
and the number of generations required to evolve a working program
increases.

For example, let us suppose that we have a simple programming language
in which there are only ten possible statements - not types of statements,
but ten statements total. Also, let us suppose that the number of
working programs increases as the square root of the total number
of possible programs. (My first system was rather different, in that
the number of possible statements increased as the number of variables
increased. But using my system as a rough guide, for a program that
is 3 statements long - the minimum needed for statement set 1 - the
density of solutions $D$ was proportional to $\frac{1}{V^{4}}$ for
large $V$, where $V$ is the number of variables. The size of the
solution space was proportional to $V^{9}$, again for large $V$.
So for large $V$, the number of working programs must be proportional
to $V^{5}$. This is slightly more than the square root of the total
number of possible programs.)

Then, in our hypothetical example, if the program is 20 statements
long, the total solution space is $10^{20}$, there are about $10^{10}$
possible working programs, and the density of working programs is
$10^{-10}$. The number of generations needed to evolve a solution
is of the order of $10^{5}$, which is quite doable. But if the problem
requires a program that is only twice as long (40 statements), there
are $10^{40}$ possible programs, only about $10^{20}$ of them work,
and the number of generations is of the order $10^{10}$. At this
point, you need either a cluster of machines, or an uninterruptible
power supply and some patience. Make the problem harder again, so
that the program needs 80 statements, and the size of the solution
space is $10^{80}$, there are about $10^{40}$ working programs,
and it will take of the order of $10^{20}$ generations to evolve
a solution. Now you need a big cluster \emph{and} a lot of patience.
Make the problem harder once more, so that the program needs 160 statements,
and the size of the solution space is $10^{160}$, there are about
$10^{80}$ working programs, and it will take of the order of $10^{40}$
generations to evolve a solution. This is hopeless - genetic programming
simply isn't a reasonable way of solving a problem of this size, on
any hardware. And this is for a program that is only 160 statements
long! As programs go, this is still a very small one. A competent
programmer can write such a program in a day or two - if the problem
is within the scope of the programmer's competence. If it is a problem
that the programmer has no idea how to approach, the literature will
help - if the answer has been published.

It seems, then, that genetic programming is best for smaller problems
\emph{that we don't yet know how to solve}. If the problem is a standard
one, like parity or sorting, a human programmer will run rings around
genetic programming. But for problems where the solution is not yet
known to mankind, genetic programming beats both brute-force search
and (at least sometimes) human ingenuity. Ironically, then, humans
are better at the boring parts of programs, and genetic programming
is better at the really interesting problems (as long as they are
small).

For program-like problems, one way to keep the problem small is to
use the most powerful statements that you can. {}``Small'' really
means that the universe of all possible programs is small. This in
turn means that only a small number of statements is needed to write
the program. Also, it seems to help if the statement set is all at
the same level of abstraction.

\section{FURTHER QUESTIONS}

What is the formula for the {}``slightly worse'' part of {}``slightly
worse than $\frac{1}{\sqrt{D}}$''? Is it $\log(\frac{1}{D})$? If
so, then we have $G=k(\frac{1}{\sqrt{D}})\log(\frac{1}{D})$. What
is the interpretation of $\frac{1}{D}$? Is it a valid measure of
the difficulty of the problem for that statement set?

What is the proportionality {}``constant''? (It's not really constant,
since it varies with statement set, population size, and maybe other
parameters.)

Perhaps the most interesting question: What is the density of working
solutions for DNA-based biological systems in the total possible DNA
space? Or, on a smaller scale, what is the density of working solutions
for protein sequences that will bind at a specific site, when implemented
in the {}``statements'' of DNA?

\end{document}